\documentclass[10pt, doublecolumn]{IEEEtran}

\usepackage[utf8]{inputenc}
\usepackage{amsmath,bm,amsthm,amsfonts,amssymb}
\usepackage{graphicx}
\usepackage{xcolor}
\usepackage{xspace}
\usepackage{hyperref}
\usepackage{mathtools}

\newtheorem{definition}{Definition}
\newtheorem{proposition}{Proposition}

\newtheorem{theorem}{Theorem}

\newcommand{\shorten}[2]{#2}       

\newcommand{\ie}{i.e.,\xspace}

\renewcommand{\paragraph}[1]{\vspace{3mm}\noindent\textbf{#1}}

\renewcommand{\b}[1]{{\bm{#1}}}   
\renewcommand{\(}{\left(}           
\renewcommand{\)}{\right)}

\newcommand{\1}{\b{1}}              
\newcommand{\0}{\b{0}}              
\newcommand{\x}{\b{x}}

\newcommand{\h}{\b{h}}

\newcommand{\e}{\b{e}}
\newcommand{\eps}{\bm{\varepsilon}} 
\renewcommand{\L}{\b{L}}            
\newcommand{\U}{\b{U}}              
\newcommand{\bSigma}{\b{\Sigma}}    
\newcommand{\bLambda}{\b{\Lambda}}  
\newcommand{\bOmega}{\b{\Omega}}  

\newcommand{\I}{\b{I}}
\newcommand{\X}{\b{X}}

\newcommand{\W}{\b{W}}              
\newcommand{\A}{\b{A}}             
             
\newcommand{\B}{\b{B}}              

\DeclareMathOperator*{\argmin}{arg\,min}

\newcommand{\LG}{\L_{\hspace{-1px}G}}              
\newcommand{\LJ}{\L_{\hspace{-1px}J}}              
\newcommand{\LT}{\L_{\hspace{-1px}T}}              

\newcommand{\UG}{\U_{\hspace{-1px}G}}              
\newcommand{\UJ}{\U_{\hspace{-1px}J}}              
\newcommand{\UT}{\U_{\hspace{-1px}T}}              

\newcommand{\GFT}[1]{\textrm{GFT}\hspace{-.0mm}\{#1\}}

\newcommand{\JFT}[1]{\textrm{JFT}\hspace{-.0mm}\{#1\}}

\newcommand{\Rbb}{\mathbb{R}}

\newcommand{\E}[1]{\mathbb{E}\left[#1\right]}        
\renewcommand{\vec}[1]{\textrm{vec}\hspace{-.5mm}\(#1\)}           
\newcommand{\diag}[1]{\textrm{diag}\hspace{-.5mm}\(#1\)}           
\newcommand{\abs}[1]{\left|#1\right|}           
\newcommand{\trace}[1]{\textrm{tr}\hspace{-.5mm}\(#1\)}            
\newcommand{\norm}[1]{\left\lVert#1\right\rVert}        
\newcommand{\hermitian}{*}                      
\newcommand{\delequal}{\overset{\Delta}{=}} 

\title{Predicting the evolution of stationary graph signals}
\author{ Andreas Loukas and Nathanael Perraudin}
\date{ }

\begin{document} 
\maketitle

\begin{abstract}
An emerging way of tackling the dimensionality issues arising in the modeling of a multivariate process is to assume that the inherent data structure can be captured by a graph. Nevertheless, though state-of-the-art graph-based methods have been successful for many learning tasks, they do not consider time-evolving signals and thus are not suitable for prediction.
Based on the recently introduced joint stationarity framework for time-vertex processes, this letter considers multivariate models that exploit the graph topology so as to facilitate the prediction. 
The resulting method yields similar accuracy to the joint (time-graph) mean-squared error estimator but at lower complexity, and outperforms purely time-based methods. 
\end{abstract}


\section{Introduction}
 
In the problem of modeling and predicting statistical processes, (wide-sense) stationarity is a helpful assumption, that allows us to learn the spectral characteristics of a process using very few samples. 
Especially for time-series prediction, learning from few samples is crucial, as one needs to estimate future values after only partially observing a single realization of the statistical process. 
This is the main reason why classical models for estimation and prediction of univariate processes, such as Wiener filters and auto-regressive moving average models (ARMA), rely on stationarity to produce predictions. 

For multivariate statistical processes, following the same methodology is often problematic, as the number of parameters to be estimated increases quadratically with the number variables, often rendering the problem intractable. 
A common way to deal with this dimensionality issue is to assume that there is an inherent structure to the process that can be captured by a graph. The graph assumption appears frequently in the machine learning and signal processing literature, and has been shown invaluable for tasks such as clustering~\cite{belkin2001laplacian,von2008consistency}, low-rank extraction~\cite{shahid2016fast}, spectral estimation~\cite{perraudin2016stationary,marques2016stationary} and semi-supervised learning~\cite{belkin2004semi,smola2003kernels}. Nevertheless, despite their promise, so far state-of-the-art graph-based methods predominantly ignore the time-dimension of data.

The objective of this paper is to identify multivariate models that exploit the graph structure inherent to the data so as to facilitate the task of prediction. 
From a statistical perspective, our model amounts to assuming stationarity not only with respect to the time-dimension, but also with respect of the graph topology~\cite{perraudin2016stationary,girault2015stationary}. The concept of \emph{joint} (time-vertex) stationarity, which we introduced in~\cite{perraudin2016towards}, was shown to facilitate regression as it enforces graph structure into the covariance. 
Unlike our previous work however, we here focus on prediction, where \emph{causality} is important as one needs to forecast the future in a timely manner. 

Concretely, we bring forth a decoupling theorem that allows us to decouple a joint multivariate process into independent univariate processes. This allows us to (\emph{i}) estimate the model parameters using traditional univariate techniques, and (\emph{ii}) to reduce further the computational complexity by combining the training stage with (an optimal) low-rank approximation of our data. The learned models are causal and can be used to provide optimal predictions (in the mean-squared error sense) at a cost that is equivalent to a constant number of matrix-vector multiplications. Finally, we show on a synthetic and a weather dataset, that our method outperforms single variable models and yields similar accuracy to the more expensive mean-squared error estimator suitable for joint stationary processes~\cite{perraudin2016towards}.


\section{Preliminaries}

Our objective is to model and predict the evolution of graph signals, \ie signals supported on the vertices $\mathcal{V} = \{ v_1, v_2, \ldots, v_N \}$ of a weighted undirected graph $G = (\mathcal{V}, \mathcal{E}, \W_G)$, with $\mathcal{E}$ the set of edges and $\W_G$ the weighted adjacency matrix.

\paragraph{Graph signal analysis.} 
The {Graph Fourier Transform} (GFT) of a graph signal $\x \in \mathbb{R}^N$ is defined as $\GFT{\x} = \UG^* \x$, where $\UG$ is the eigenvector matrix of the (combinatorial\footnote{Though we use the combinatorial Laplacian in our presentation, our results are applicable to any positive semi-definite matrix representation of a graph or to the recently introduced shift operator~\cite{sandryhaila2013discrete}.}) Laplacian matrix $\LG = \diag{\W_G \1_N} - \W_G = \UG \bLambda_G \UG^*$.
The GFT allows us to extend filtering to graphs~\cite{hammond2011wavelets,shuman2013emerging,shuman2013vertex}. Filtering a signal $\x$ with a graph filter $h(\LG)$ corresponds to element-wise multiplication in the spectral domain 
\begin{equation*}
    h(\LG)  \x \delequal \UG h(\bLambda_G) \UG^\hermitian \, \x,
\end{equation*}
where the scalar function $h : \mathbb{R}_+ \mapsto \mathbb{R}$, referred to as the graph frequency response, has been applied to each diagonal entry of $\bLambda_G$. 

\paragraph{Time-varying graph signals.} Let $\x_t$ be a graph signal is sampled at $T$ successive regular intervals of unit length. We refer to matrix $\X = \left[ \x_1, \x_2, \ldots, \x_T \right] \in \mathbb{R}^{N\times T}$ as a time-varying graph signal
, and $\x = \vec{\X}$ (without subscript) is the vectorized form of $\X$. 
The joint (time-vertex) Fourier transform (JFT) of $\X$ is 
\begin{align}
    \JFT{\X} = \UG^* \X \bar{\U}_T,
\end{align}
where, once more, $\UG$ is the graph Laplacian eigenvector matrix, whereas $\bar{\U}_T$ is the complex conjugate of the DFT matrix. 
In fact, $\UT$ is the eigenvector matrix of the lag operator $\LT = \UT \, \bLambda_T \, \UT^*$. Denote by $\bOmega$ the diagonal matrix of angular frequencies (\ie $\bOmega_{tt} = \omega_t = 2 \pi t/T$), we have  $\bLambda_T = e^{-j \bOmega}$, where $j = \sqrt{-1}$. 
Expressed in vector form, the joint Fourier transform becomes $\JFT{\x} = \UJ^* \x$, where $\UJ = \UT \otimes \UG $ is unitary, and operator $(\otimes)$ denotes the kroneker product. 
A \emph{joint filter} $h(\LJ)$ is a function defined in the joint spectral domain $h: \Rbb_+ \times \Rbb \mapsto \Rbb$ that is evaluated at the graph eigenvalues $\lambda$ and the angular frequencies $\omega$. The output of a joint filter is 
\begin{equation}  \label{eq:def_joint_filtering}
    h(\LJ) \x \delequal \UJ \, h(\bLambda_G,\bOmega) \, \UJ^* \x, 
\end{equation} 
and where $h(\bLambda_G,\bOmega)$ is a $NT\times NT$ diagonal matrix with  $[h(\bLambda_G,\bOmega)]_{k,k} = h(\lambda_n,\omega_\tau)$ and $k = N(\tau-1)+n$.  
For an in-depth discussion of JFT and its properties, we refer the reader to the  work by Loukas and Foucard~\cite{loukas2016frequency}.

\section{Modeling jointly stationary signals}

The first step in predicting the evolution of a signal is to choose a good model for it. 
Motivated by the importance of stationarity for modeling statistical processes, our recent work generalized stationarity to time-varying graph signals~\cite{perraudin2016towards}. The following is a variant of the definition presented in~\cite{perraudin2016stationary}. 

\begin{definition}[JWSS process]
A process $\x = \vec{\X}$ is called Jointly (or time-vertex) Wide-Sense Stationary (JWSS), if and only if (\textit{i}) $\LJ \E{\x} = \0_{NT}$, and (\textit{ii}) its covariance matrix $\bSigma_\x$ is diagonalizable by the joint Fourier basis $\UJ$. The JFT of the autocorrelation function of process $\x$ is referred to Joint Power Spectral Density (JPSD). 
\end{definition}

This definition is a generalization of the classical notion of stationarity, where now one assumes simultaneously wide-sense stationarity w.r.t. both the \emph{time} and \emph{vertex} domains. Indeed, for the combinatorial Laplacian, the first condition is equivalent to the first moment being constant. The second condition is a generalization of the invariance w.r.t. translation using the localization operator. Joint stationarity assumes the covariance to be driven by a joint filter $h(\lambda,\omega)$, referred to as JPSD, that encodes the time and vertex relations in the signal.
For an in depth discussion of the above definition, we encourage the reader to read the original publications~\cite{perraudin2016stationary,perraudin2016towards}.

\vspace{2mm}
The question we will address is: ``\textit{how to construct models that concisely capture the characteristics of a jointly stationarity process, in order to facilitate short-term prediction?}''. 

\paragraph{Joint non-causal models.} Since the covariance of a jointly stationary signal $\x$ is diagonalizable by $\UJ$, without loss of generality, we can constrain ourselves to models of $\x$ which take the form 
\begin{align}
    a(\LJ) \, \x = b(\LJ) \, \eps,
    \label{eq:GM}
\end{align}
where the innovation vector $\eps = \vec{\b{E}}$ is a random vector of some arbitrary distribution, with mean in the null space of $\LJ$ and identity covariance matrix $\bSigma_{\eps} = \I$, implying that the above model is equivalent to the one considered in our previous work~\cite{perraudin2016towards}. Moreover, matrices $a(\LJ)$ and $b(\LJ)$ are arbitrary joint filters (and not necessarily polynomials). 
According to proposition~\ref{prop:general_model}, model \eqref{eq:GM} is natural, as the space of JWSS signals is exactly equal to that of signals generated by it.

\begin{proposition}
    Process $\X$ is JWSS iff it is generated by \eqref{eq:GM}.  
    \label{prop:general_model}
\end{proposition}

\shorten{(The proof will appear in the long version of the paper.)}{
\begin{proof}
    To establish ``\textit{if}'' direction, we have to show that the two conditions of joint stationarity are always met for any output of the model. Let $g(\LJ) = a(\LJ)^{-1} b(\LJ)$ such that $\x = g(\LJ) \eps$, then 
\begin{align*}
    \E{ \x } = \E{ a(\LJ)^{-1} b(\LJ) \eps} =  g(\LJ) \E{\eps}.
\end{align*}
It is easy to confirm that $\LJ \E{\x} = \LJ g(\LJ) \E{\eps} = \0$, since, if $\E{\eps}$ lies in the null-space of $\LJ$, then so does $g(\LJ) \E{\eps}$.
The covariance of $\x$ is  
\begin{align*}
    \bSigma_\x &= g(\LJ) \bSigma_{\eps} g(\LJ)^\hermitian = g(\LJ) g(\LJ)^\hermitian,
\end{align*}
which is diagonalizable by $\UJ$. 
To prove the ``\textit{only if}'' direction, we observe that every JWSS signal with mean $\bar{\x}$ and JPSD $h$ can be produced by the model by setting $\E{\eps} = \bar{\x}$, and $g(\LJ) g(\LJ)^\hermitian = h(\LJ)$ (this is always possible as covariance matrices are positive semi-definite).    
\end{proof}
} 

\paragraph{Joint causal models.} Yet, despite its generality, model~\eqref{eq:GM} is not always causal. This is problematic for the task of prediction, where one needs to forecast the future in a timely manner. It is more practical to assume that the output at time $t$ can be expressed as a function of the input-output variables at previous timesteps, yielding the joint causal model
\begin{align}
\sum_{p = 0}^P a_p(\LG) \, \x_{t-p} &=  \sum_{q = 0}^Q b_{q}(\LG) \, \eps_{t-q}   
\label{eq:joint_causal_model}
\end{align}
for some model orders $P$ and $Q$, and with $\eps_t$ being the $t$-th column of matrix $\b{E}$. Moreover, for the canonical form, we set $a_0(\LG) = b_0(\LG) = \I$.  Obviously, every causal model can be written in this form for $P\rightarrow \infty$ and $Q \rightarrow \infty$. For this reason, we refer to processes generated by the above model as {joint causal processes}. 
\section{Predicting evolving graph signals}

Our objective is to forecast to the evolution of an observed jointly stationary process $\x$ with zero mean and JPSD $h(\omega,\lambda)$. To start with, we will suppose that $\x$ is a joint causal process, with known parameters $a_p(\lambda)$ and $b_q(\lambda)$ and we will derive an optimal predictor for the future values of the process. In the second part, we show how to estimate a joint causal model that approximates an arbitrary process (not necessarily causal).  Finally, we demonstrate how to reduce the computational complexity of model estimation.

\paragraph{Prediction.} Suppose that $\x$ is the output of a joint causal model, where the input $\eps$ has zero mean and identity covariance. This implies that $\x$ abides to 
\begin{align}
    \x_{t} &=  \sum_{q = 0}^Q b_q(\LG)\, \eps_{t-q} - \sum_{p = 1}^P a_p(\LG) \,\x_{t-p}.   
\label{eq:jARMA1}
\end{align}
Let the subscript $\x_{t | t-1}$ denote that the random vector $\x_t$ is conditioned on the (already observed) vectors $\{\x_1, \ldots, \x_{t-1}\}$. 
We predict vector $\x_{t}$, based on $\x_1, \x_2, \ldots, \x_{t-1}$, thus obtaining the \emph{one-step predictor} $\tilde{\x}_{t}$, using the conditional expectation 
\begin{align}
    \tilde{\x}_{t} &= \E{ \x_{t|t-1} } \notag \\
    &\hspace{-3mm}= \sum_{q = 0}^Q b_q(\LG) \E{\eps_{t-q|t-1}} - \sum_{p = 1}^P a_p(\LG)\, \x_{t-p} \notag \\
    &\hspace{-3mm}= \E{\eps_{t}} + \sum_{q = 1}^Q b_q(\LG) \( \x_{t-q|t-1} - \tilde{\x}_{t-q} \) - \sum_{p = 1}^P a_p(\LG) \x_{t-p} \notag \\
    &\hspace{-3mm}= \sum_{q = 1}^Q b_q(\LG) \( \x_{t-q} - \tilde{\x}_{t-q}  \) - \sum_{p = 1}^P a_p(\LG)\, \x_{t-p}, 
    \label{eq:predictor}
\end{align} 
where above we have exploited the fact that $\E{\eps_t} = 0$. Notice that, in the third step we compute the unobserved input vectors $\eps_{t-q|t-1}$ as the difference between the observations and the predictions in previous time-steps. In the following, we will show that this corresponds to the best possible choice, as it yields the minimum possible mean-squared error. %
We obtain a $k$-step predictor by repeating the above computation $k$ times.   

One maybe tempted to apply a similar procedure in order to predict unknown values along the graph dimension. The model above however does not render itself suitable for such a task, even when functions $a_p$ and $b_q$ are polynomials (and thus locally computable in the graph). In contrast to prediction along the time domain, in purely graph prediction (or inpainting) the value of a node is dependent on all of its neighbors. Therefore, node values cannot be predicted in isolation, but have to be solved jointly. 

\paragraph{Error analysis.} The one-step prediction error is $\e_t = \x_t - \tilde{\x}_{t}$. Similar to the classical case~\cite{ljung1998system}, $\e_t$ depends only on the unknown innovations $\eps_t$ and the mean-squared error achieved is the smallest possible.
To see this, we first need to show that $\e_t = \eps_t$, or equivalently that $ \b{d}_t = \e_t - \eps_t = 0$. Expanding~\eqref{eq:predictor} in the definition of $ \b{d}_t$, we have that 
\begin{align}
    \b{d}_t = \e_t - \eps_t = \x_t - \tilde{\x}_{t} - \eps_t & = \sum_{q = 1}^Q b_q(\LG) \( \eps_{t-q} - \e_{t-q} \)  \notag \\
    & = - \sum_{q = 1}^Q b_q(\LG) \b{d}_{t-q}.
\end{align} 
Therefore, $\sum_{q = 0}^Q b_q(\LG) \b{d}_{t-q} = 0$ for every $t$, which, under the assumption that the noise model is invertible\footnote{System $\sum_{q = 0}^Q b_q(\LG) \b{d}_{t-q} = 0$ has exactly one solution when matrix $b_0(\LG) \oplus b_1(\LG) \oplus \ldots \oplus b_Q(\LG) $ is invertible, or equivalently when $b_q(\LG)$ is invertible for each $q$.}, implies $\b{d}_t = 0$.
Directly, we find that $\e_t = \eps_t$ and the one-step mean-squared error is equal to 
\begin{align}
    \frac{\E{\norm{ \x_t - \tilde{\x}_{t} }_2^2}}{N} = \frac{\E{\eps_t^\hermitian \eps_t}}{N} = \frac{\trace{\E{\eps_t \eps_t^\hermitian}}}{N} = \frac{\trace{\bSigma_{\eps_t}}}{N}.
\end{align}
Since $\eps_t$ is unknown at time $t$, the above corresponds to the smallest achievable mean-squared error.

\paragraph{Model estimation.} Suppose that we want to identify the parameters of a joint causal model (\ie functions $a_p$ and $b_q$ for every $p$ and $q$) which best match an observed process $\X \in \mathbb{R}^{N \times T}$. The canonical way to achieve this would be to minimize the prediction error residuals by solving the following (non-linear) system of $N \times T$ equations involving $(P + Q) \, N$ unknowns\footnote{A matrix function of an $N \times N$ matrix has $N$ degrees of freedom.}
\begin{align}
    \min_{a_p, b_q} \norm{ \x_{t+1} - \tilde{\x}_{t+1}(a_p, b_q) }^2_2, 
\end{align}
where by $\tilde{\x}_{t+1}(a_p, b_q)$ we refer to the causal model based on $a_0(\LG), \ldots, a_P(\LG)$ and $b_0(\LG), \ldots, b_Q(\LG)$. 
In the following, we will use the \emph{decoupling theorem} in order to simplify this problem by splitting it to a number of independent and well-studied problems with smaller complexity.

\begin{theorem}[Decoupling theorem]
Let $\eps_t$ and $\x_t$ abide to model~\eqref{eq:joint_causal_model} 
\shorten{and set 
}{
\begin{align}
    \sum_{p = 0}^P a_p(\LG) \x_{t-p}  = \sum_{q = 0}^Q b_q(\LG) \eps_{t-q},
    \label{eq:GM_time}
\end{align}
where $\LG$ is a symmetric diagonalizable $N \times N$ matrix written as $\LG= \U \b{D} \U^{\hermitian}$. Set }
$\hat{\varepsilon}_t{(n)} = \(\UG^{\hermitian} \eps_{t}\)(n)$ and $\hat{x}_t{(n)} = \(\UG^{\hermitian} \x_{t}\)(n)$. 
Up to an isometric rotation by $\UG$, the input-output relation of $\hat{\varepsilon}_t{(n)}$ and $\hat{x}_t{(n)}$ for every $n$ is given by an ARMA(P,Q) model $\sum_{p = 0}^P a_p(n) \hat{x}_{t-p}(n)  = \sum_{q = 0}^Q b_q(n) \, \hat{\varepsilon}_{t-q}(n)$, with $a_p(n)$ and $b_q(n)$ scalars. 
%
%
\label{theorem:separability}
\end{theorem}
\shorten{(The proof will appear in the long version of the paper.)}{
\begin{proof}
   The $n$-th element of the left part of~\eqref{eq:GM_time} rotated by unitary matrix $\UG^{\hermitian}$ is
    \begin{align} 
    \(\UG^{\hermitian} \sum_{p = 0}^P a_p(\LG) \x_{t-p}\) (n) &= \(\sum_{p = 0}^P a_p(\LG) \UG^{\hermitian} \x_{t-p}\) (n) \notag \\
    &=  \sum_{p = 0}^P a_p(n) \hat{x}_{t-p}(n), 
    \label{eq:ll1}
    \end{align}
    with $a_p(n) = [a_p(\LG)]_{nn}$.
    Similarly, setting $b_q(n) = [b_q(\LG)]_{nn}$ we have that
    \begin{align} 
        \(\UG^{\hermitian} \sum_{q = 0}^Q b_q(\LG) \eps_{t-q}\) (n) &= \(\sum_{q = 0}^P b_q(\LG) \UG^{\hermitian} \eps_{t-q}\) (n) \notag \\
        &=  \sum_{q = 0}^Q b_q(n) \hat{\eps}_{t-q}(n). 
        \label{eq:ll2}
    \end{align}
    Combining expressions~\eqref{eq:ll1}~and~\eqref{eq:ll2}, the decoupling theorem is concluded. 
\end{proof}
}
Therefore, under the appropriate isometric rotation, the joint causal model decomposes into $N$ independent ARMA models, 
with two important consequences: (\textit{i}) in the graph frequency domain, the joint causal model estimation problem can be split to $N$ independent problems, each involving $T$ equations and $P + Q$ unknowns. 
(\textit{ii}) Despite being non-linear, each of the $N$ problems we now have to solve corresponds to fitting an ARMA to a time-series. We can therefore use a number of well understood methods to solve it, such as the subspace Gauss-Newton approach of Wills and Ninness~\cite{wills2008gradient}.       
The exact coefficients of the joint causal model are then found by an inverse graph Fourier transform. We note that, especially for small to medium sized graphs, the cost of eigenvalue decomposition (inherent in joint models), is overshadowed by that of model estimation.

\paragraph{Low-rank models.} The computational overhead of model estimation can be reduced by purposely ignoring a subset of the data. For instance, one may consider only a subset of the time-series and ignore the rest, saving considerably in terms computation complexity. Nevertheless, such a simplistic scheme will incur significant error when the variance of the data is distributed evenly among all time-series. To capture more concisely the variance inherent the data, we can instead perform the selection process after first rotating the data. Concretely, denote by $\mathcal{S}$ an index set of size $K = |\mathcal{S}|$ and let $\U$ be a unitary (rotation) matrix. 
The $\{\U, \mathcal{S}\}$ low-rank approximation of $\X$ is \shorten{$\tilde{\X}_{\U\hspace{-0.5mm}, \mathcal{S}} = \U \, \I_{\mathcal{S}} \, \U^\hermitian \X$,}{$$\tilde{\X}_{\U\hspace{-0.5mm}, \mathcal{S}} = \U \, \I_{\mathcal{S}} \, \U^\hermitian \X,$$} 
where the diagonal indicator matrix $\I_{\mathcal{S}}$ has $[\I_\mathcal{S}]_{ii} = 1$ if $i \in \mathcal{S}$, and zero otherwise. For prediction, which is an online task, the low-rank approximation needs to be performed solely w.r.t the graph dimension. For offline regression tasks, a low-rank approximation scheme could also consider the time-dimension.

According to Theorem~\ref{theorem:low-rank}, if we are interested in the expected behavior, the \textit{best low-rank approximation} of a JWSS process uses the eigenvectors of the graph $\UG$ to rotate the data (corresponding to a GFT). Nevertheless, by the decoupling theorem, rotation by $\UG$ also decouples the time-series, and is the first step of model estimation. Therefore, the low-rank approximation can be effortlessly combined with model estimation, by only modeling the time-series in the decoupled space specified by set $\mathcal{S}$. Using this scheme, we attain a complexity reduction by a factor of $N/K$ for model estimation.

\begin{theorem}
    Let $\X$ be a zero-mean JWSS process. The best rank-$K$ approximation of $\X$ is given by 
    \begin{align}
       \{\U_G, \mathcal{S}^\star\} = \argmin_{\U, \mathcal{S}} \E{\norm{ \X - \tilde{\X}_{\U, \mathcal{S}} }_F^2} \quad s.t. \quad |\mathcal{S}| = K \notag 
    \end{align}
    where $\mathcal{S}^\star$ contains the indices of the top-$K$ diagonal elements of $\U_G^{\hermitian} \, \E{\X \X^\hermitian} \U_G$. 
    \label{theorem:low-rank}
\end{theorem}
\shorten{(The proof of the Theorem, which is inspired by the Eckart–Young–Mirsky theorem~\cite{eckart1936approximation,markovsky2008structured}, will appear on the long version of this paper.)}{
\vspace{-3mm}
\begin{proof}
	The proof is inspired by the Eckart–Young–Mirsky theorem~\cite{eckart1936approximation,markovsky2008structured}.
    Let us define $\A = \U (\I -  \I_\mathcal{S}) \U^{\hermitian}$. Since $\U$ is unitary the expected approximation error is equivalent to
    \begin{align}
        \E{\norm{ \X - \tilde{\X} }_F^2} &= \E{\norm{ \A \X }_F^2} 
        = \trace{\A \, \E{\X \X^\hermitian} \A^\hermitian  }. \notag 
    \end{align}
    By Theorem 2 of~\cite{perraudin2016towards}, for each time instant $\x_t$, the graph signal is stationary with covariance $\bSigma_t = \E{\x_t \x_t^\hermitian}=\UG g_t^2(\bLambda_G)\UG^{\hermitian}$, which implies that 
	\begin{eqnarray*}
	\bSigma_G & = & \E{\X \X^{*}}
		=  \E{\sum_{t=1}^T\x_t\x_t^*} 
	  = \sum_{t=1}^T \bSigma_{t}\\
	 & = & \UG \left(\sum_{t=1}^T g_{t}^2(\bLambda_G)\right) \UG^{*}
	  = \UG g^2(\bLambda_G)\UG^{*},
	\end{eqnarray*}
    for $g(\lambda) = \sqrt{\sum_{t=1}^T g_t^2(\lambda)}$. 
    As a result, we observe that $\bSigma_G = \E{\X \X^\hermitian}$ is an $N\times N$ matrix jointly diagonalizable with the graph Laplacian $\LG$. 
    Let us order $g(\bLambda_G)$ such that $g(\lambda_1) \geq \ldots \geq g(\lambda_n)$. 
    We can write
    \begin{align}
        \E{\norm{ \X - \tilde{\X} }_F^2} &= \trace{\A \, \bSigma_G \, \A^\hermitian } \notag\\
        &\hspace{-10mm}= \norm{ \bSigma_G^{1/2} - \U \I_\mathcal{S} \U^{\hermitian} \bSigma_G^{1/2} }_F^2 \notag \\
        &\hspace{-10mm}= \norm{ g(\bLambda_G) - \UG^\hermitian \U \I_\mathcal{S} \U^{\hermitian} \UG g(\bLambda_G) }_F^2.
    \end{align}
     Setting $\B = \U_G^\hermitian \U \I_\mathcal{S} \U^{\hermitian} \U_G$, the above expression becomes
     \begin{align}
         \E{\norm{ \X - \tilde{\X} }_F^2} &= \sum_{i = 1}^N \abs{ g(\lambda_i) - \B_{ii} g(\lambda_i) }^2 + \sum_{i \neq j} \abs{ \B_{ij} g(\lambda_i) }^2 \notag \\
         &\hspace{-12mm}\geq  \sum_{i = 1}^N \abs{ g(\lambda_i) }^2 \abs{ 1 - \B_{ii} }^2 \notag \\
         &\hspace{-12mm}\geq \sum_{i = K+1}^N \abs{ g(\lambda_i) }^2  = \E{\norm{\X -\tilde{\X}_{\UG,\mathcal{S}^*}}_F^2},
     \end{align}
     where in the third step, we use the fact that $\B_{ii}$ is bounded by $1$ and can be $1$ at most $K$ times. The last expression shows that a global minimum is achieved for $\tilde{\X}_{\UG,\mathcal{S}^*}$, where $\mathcal{S}$ contains the largest $K$ components of function $g(\lambda)$.
\end{proof}
}

\section{Evaluation}

We compare the joint causal models proposed in this paper, with (\textit{i}) \emph{disjoint causal} models, which model and predict each time-series independently (using $N$ ARMA models), and (\textit{ii}) with \emph{joint non-causal} models, which consider all relationships between variables. A disjoint model (which corresponds to setting $\L_G = \I$) effectively ignores the graph topology and models each time-series as a causal WSS (wide-sense stationary in time) process. This is a valid approach as any JWSS process also WSS (see Theorem 2 in~\cite{perraudin2016towards}). On the other hand, prediction using a non-causal joint model corresponds to the minimum mean-squared error estimator in the general case, and to a MAP estimator in case the process is Gaussian distributed~\cite{perraudin2016towards}. Thus, if the JPSD is correctly estimated, non-causal joint predictors are optimal---though at the expense of higher computation per prediction and space (for storing the JPSD), as compared to causal methods. 

In our experiments, we split the data (along the time-dimension) in two halfs and used the first for model estimation. Then, for each time $t = T/2 + 1, \ldots, T$ we computed the relative $k$-step prediction error $\norm{\tilde{\x}_{t+k|t} - \x_{t+k}}_2 / \norm{\x_{t+k}}_2$. We report the median prediction error and use errorbar to indicate one standard deviation. We also insert a small horizontal offset to improve visibility. Though the causal models (joint and disjoint) estimated in our experiments were not sensitive to the selected order, in the following we illustrate prediction only for the best orders (always below 3), identified using exhaustive search. 

\begin{figure}[t]
\hspace{-3mm}\includegraphics[width=1.1\columnwidth]{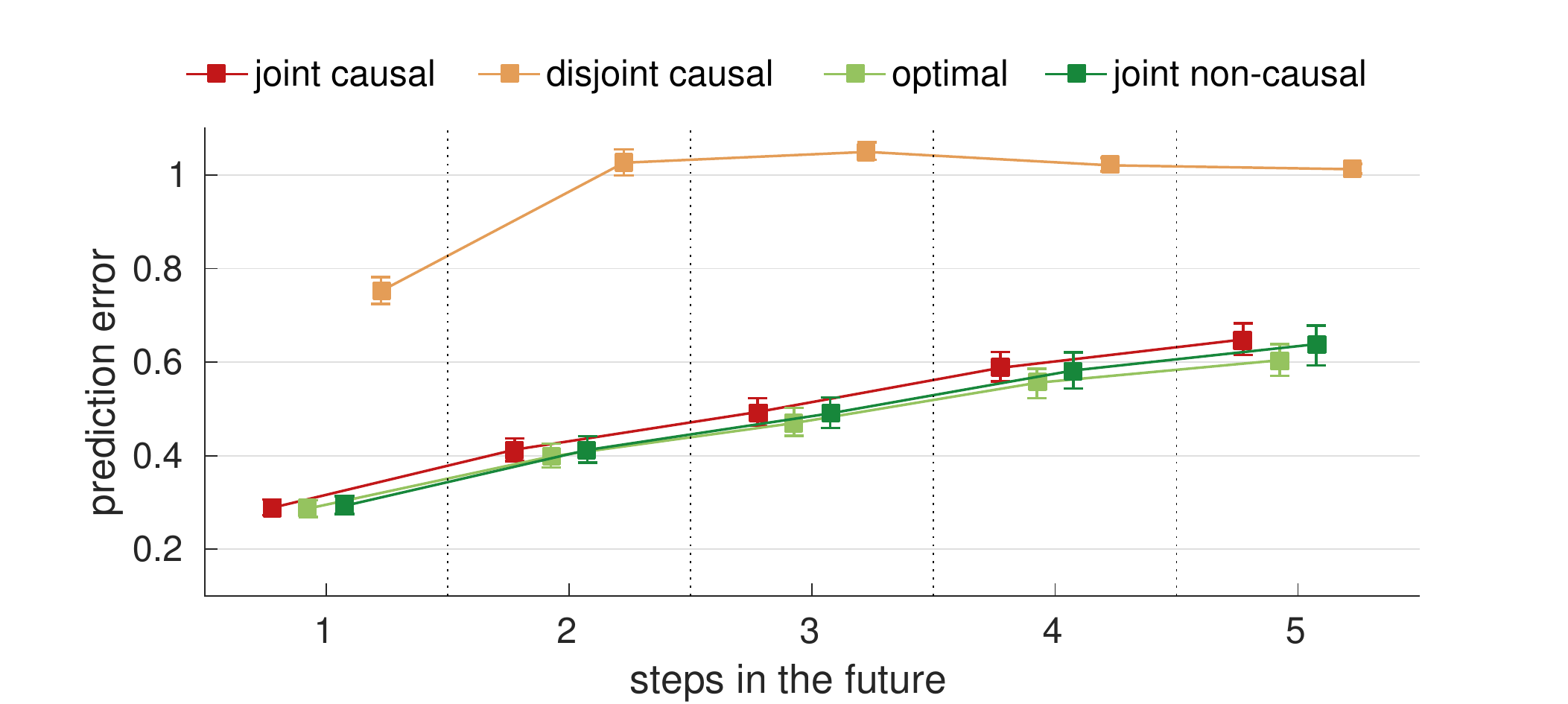}\vspace{-0mm}
\caption{Joint models predict very closely the evolution of a wave on the graph. Due to the non-separable form of the JPSD, a purely time-based prediction method (disjoint causal model) yields poor accuracy.\vspace{-0mm}}
\label{fig:wave}
\end{figure}

\paragraph{Wave equation.} We simulated $T = 200$ steps of a wave equation~\cite{grassi2016tracking} under white Gaussian input on a random geometric graph of $N = 50$ nodes and average degree 5. The scenario was motivated by the fact that the wave partial different equation has a closed form solution that corresponds to a non-separable \footnote{A JPSD is separable if it can be written as $h(\lambda,\omega) = h_G(\lambda) \cdot \h_T(\omega)$, implying an independence of the time and graph dimensions} and causal JPSD. Our experiments were repeated 10 times. As shown in Fig.~\ref{fig:wave}, the tight dependence between temporal and graph frequency contents present in a wave render prediction using a disjoint causal model inaccurate (by ignoring the graph dimension, a purely time-based method cannot capture the strong correlations between variables). On the other hand, both joint methods closely approximate the optimal solution (corresponding to a MAP estimator that uses the ground truth covariance).

\begin{figure}[t]
\hspace{-3mm}\includegraphics[width=1.1\columnwidth]{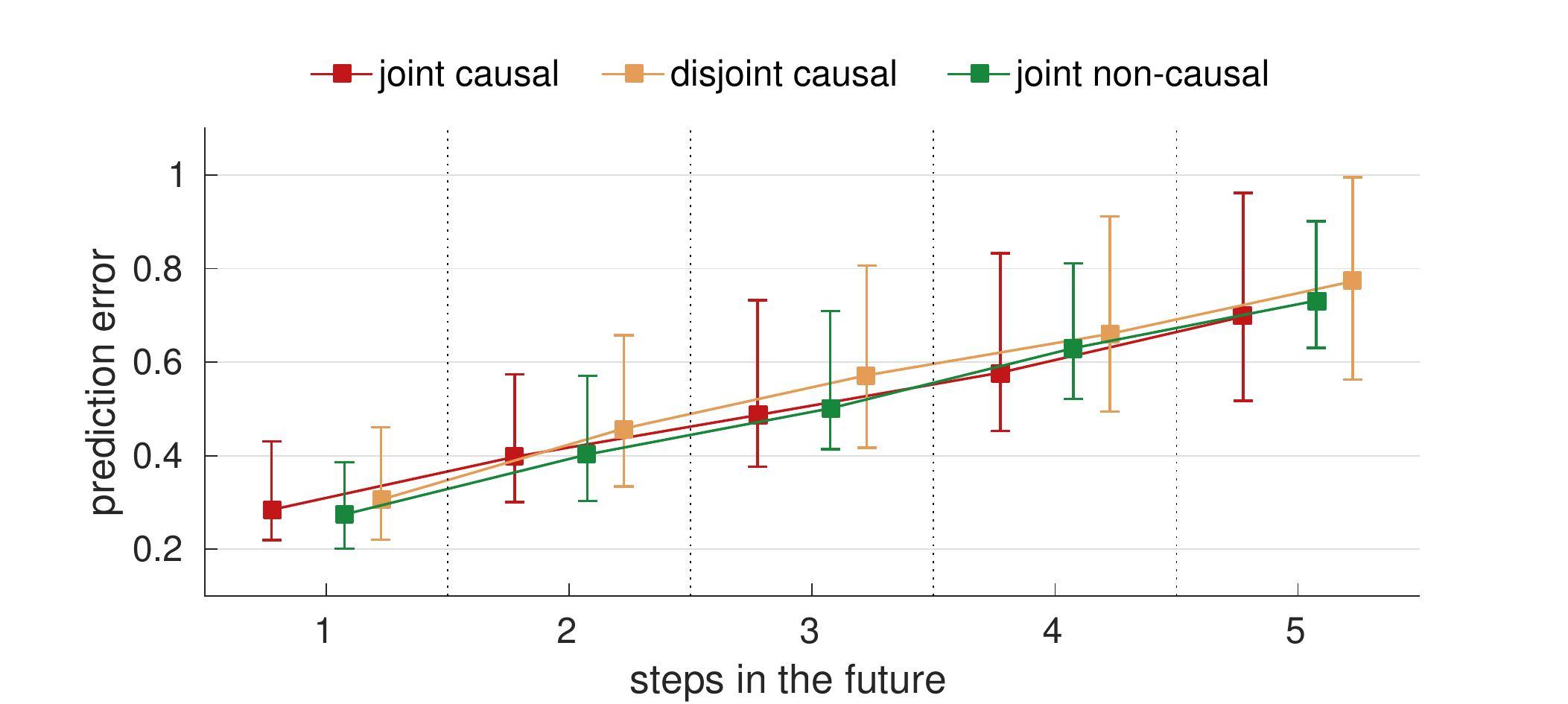}\vspace{-0mm}
\caption{The joint causal and non-causal models achieved the best prediction accuracy in our weather dataset, with a median prediction of approx. 0.4 for a 2-hour horizon (steps=2). \vspace{-0mm}}
\label{fig:molene_error}
\end{figure}

\paragraph{Weather dataset.} The second experiment used a weather dataset depicting the temperature of $N=32$ weather stations in the region of Brest (France), over a span of $T = 14\times24$ hours\footnote{Access to the raw data is possible directly from \url{https://donneespubliques.meteofrance.fr/donnees_libres/Hackathon/RADOMEH.tar.gz}}. The graph used was a 3-nearest neighbor graph built from the coordinates of the weather stations. Fig.~\ref{fig:molene_error} depicts the prediction error for 1 to 5 steps (hours) in the future. Since in this case the JPSD is (almost) separable, the disjoint causal predictor gives relevant predictions. Nevertheless, it is always outperformed by joint models. Note that the optimal predictor is not illustrated, since the ground truth JPSD of the data is unknown. We remark that the joint non-causal method has been shown to outperform non-stationarity based methods for prediction (such as TV and Tikhonov extrapolation)~\cite{perraudin2016stationary}.  

\begin{figure}[t]
\hspace{-3mm}\includegraphics[width=1.1\columnwidth]{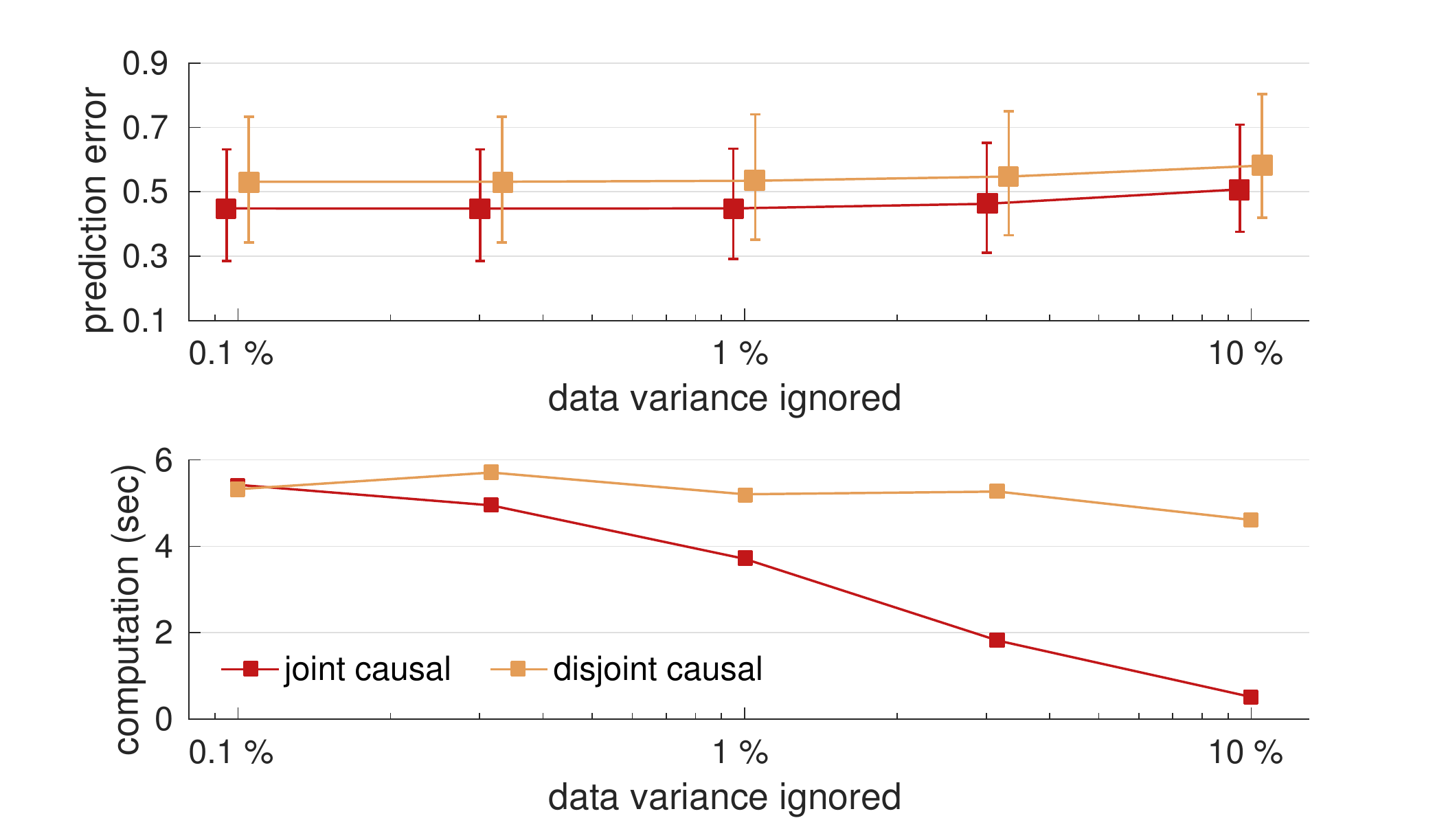}\vspace{-0mm}
\caption{At the expense of a small decrease in accuracy --here measured by the percentage of the data variance ignored-- the estimation of joint causal models becomes very scalable.\vspace{-0mm}}
\label{fig:molene_computation}
\end{figure}

Lastly, Fig.~\ref{fig:molene_computation} examines the effect of low-rank prediction to the 2-steps prediction error and model estimation time for the two causal models. In this experiment, we tested low-rank predictors which ignored a given percentage of the data variance (as estimated by the training data). Moreover, we considered the full dataset of $T = 31\times 24$ hours. As supported by our theoretical results, performing the approximation in the graph spectral domain (joint causal) far outperforms a naive approximation in the native graph domain (disjoint causal), yielding a significant computational benefit at the expense of only a small decrease prediction accuracy.

\section{Conclusion}
While, this contribution focuses on processes living on graphs, the proposed method can be applied to traditional multivariate random processes, by connecting them with a nearest neighbors graphs. In this case, compared to classical multivariate ARMA models, the proposed method requires less training data and a smaller amount of computation because the number of parameters to be estimated is low (less than $NT$). A more detailed comparison is left for future work. Note also that the current method is limited by the graph Fourier transform that require the diagonalization of the graph Laplacian $\LG$. How to scale further than a few thousands nodes is still an open question that requires to a clever replacement of the decoupling theorem. Solving this issue would allow us to model user behaviors on large graphs such as Wikipedia.

\bibliographystyle{plain}
\bibliography{references}
\end{document}